\def\year{2020}\relax
\documentclass[letterpaper]{article} 
\usepackage{aaai20}  
\usepackage{times}  
\usepackage{helvet} 
\usepackage{courier}  
\usepackage[hyphens]{url}  
\usepackage{graphicx} 
\usepackage{graphicx}  
\usepackage[table,xcdraw]{xcolor}
\usepackage{tikz}
\usepackage{subcaption}
\usepackage{multirow}
\usepackage{algorithm}
\usepackage{algorithmicx}
\usepackage{algpseudocode}
\usepackage{booktabs}
\newcommand{\citet}[1]{\citeauthor{#1}~\shortcite{#1}}
\title{Learning Sparse Sharing Architectures for Multiple Tasks}
\author{Tianxiang Sun,\thanks{Equal contribution.} Yunfan Shao,\footnotemark[1]  Xiaonan Li, Pengfei Liu, Hang Yan, Xipeng Qiu,\thanks{Corresponding author: Xipeng Qiu (xpqiu@fudan.edu.cn).} Xuanjing Huang\\
Shanghai Key Laboratory of Intelligent Information Processing, Fudan University\\
School of Computer Science, Fudan University\\
\{txsun19, yfshao19, pfliu14, hyan19, xpqiu, xjhuang\}@fudan.edu.cn, lixiaonan@stu.xidian.edu.cn 
}

\begin{document}

\maketitle

\begin{abstract}
Most existing deep multi-task learning models are based on parameter sharing, such as hard sharing, hierarchical sharing, and soft sharing. How choosing a suitable sharing mechanism depends on the relations among the tasks, which is not easy since it is difficult to understand the underlying shared factors among these tasks. In this paper, we propose a novel parameter sharing mechanism, named \emph{Sparse Sharing}. Given multiple tasks, our approach automatically finds a sparse sharing structure. We start with an over-parameterized base network, from which each task extracts a subnetwork. The subnetworks of multiple tasks are partially overlapped and trained in parallel. We show that both hard sharing and hierarchical sharing can be formulated as particular instances of the sparse sharing framework. We conduct extensive experiments on three sequence labeling tasks. Compared with single-task models and three typical multi-task learning baselines, our proposed approach achieves consistent improvement while requiring fewer parameters.
\end{abstract}

\section{Introduction}

\noindent Deep multi-task learning models have achieved great success in computer vision \cite{DBLP:conf/cvpr/MisraSGH16,DBLP:conf/cvpr/ZamirSSGMS18} and natural language processing \cite{DBLP:conf/icml/CollobertW08,DBLP:journals/corr/LuongLSVK15,DBLP:conf/acl/LiuQH17}. Multi-task learning utilizes task-specific knowledge contained in the training signals of related tasks to improve performance and generalization for each task \cite{DBLP:journals/ml/Caruana97}.

\begin{figure}[htb]
	\centering
	\begin{subfigure}{0.45\linewidth}
        \centering
        \includegraphics[scale=.36]{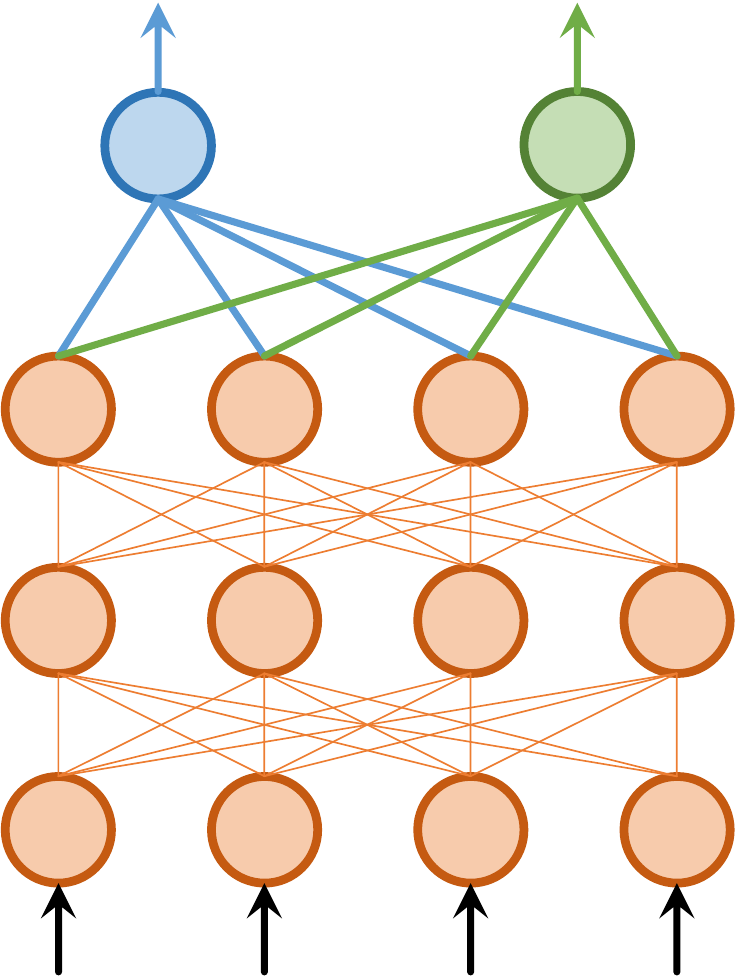}
        \caption{Hard sharing}
        \label{sfig:hard-sharing}
        \end{subfigure}\hfill
    \begin{subfigure}{0.45\linewidth}
        \centering
        \includegraphics[scale=.36]{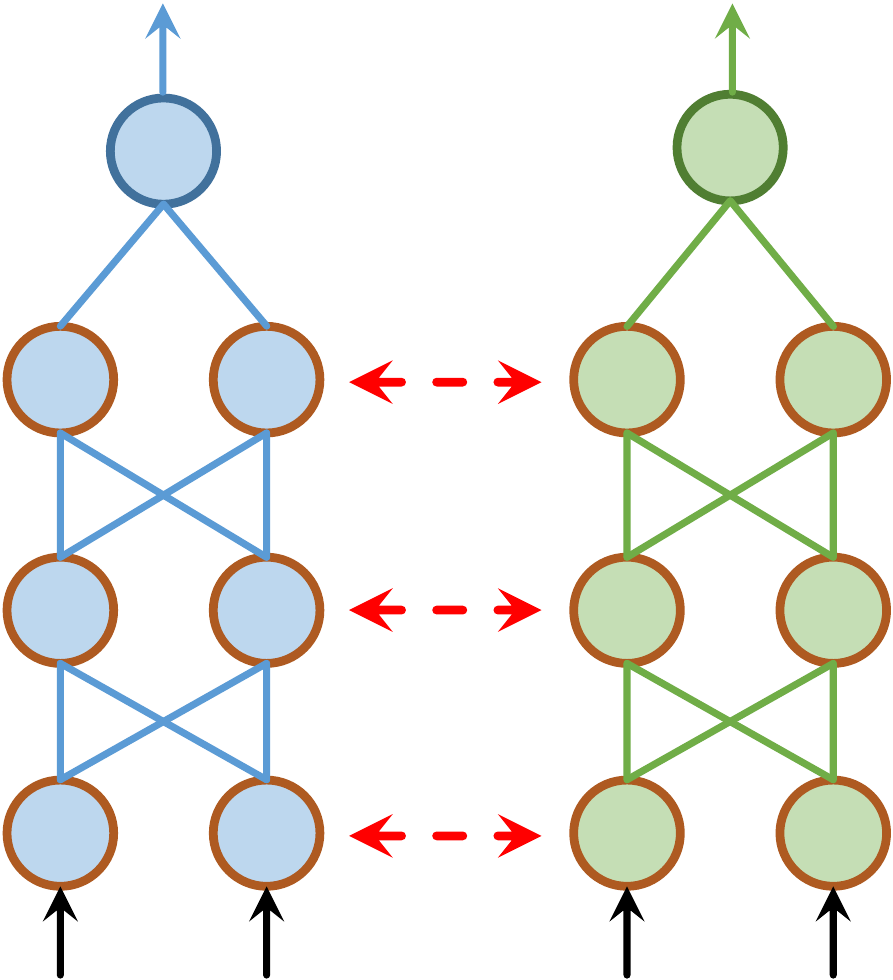}
        \caption{Soft sharing}
        \label{sfig:soft-sharing}
        \end{subfigure}\hfill
    \quad
    \begin{subfigure}{0.45\linewidth}
        \centering
        \includegraphics[scale=.36]{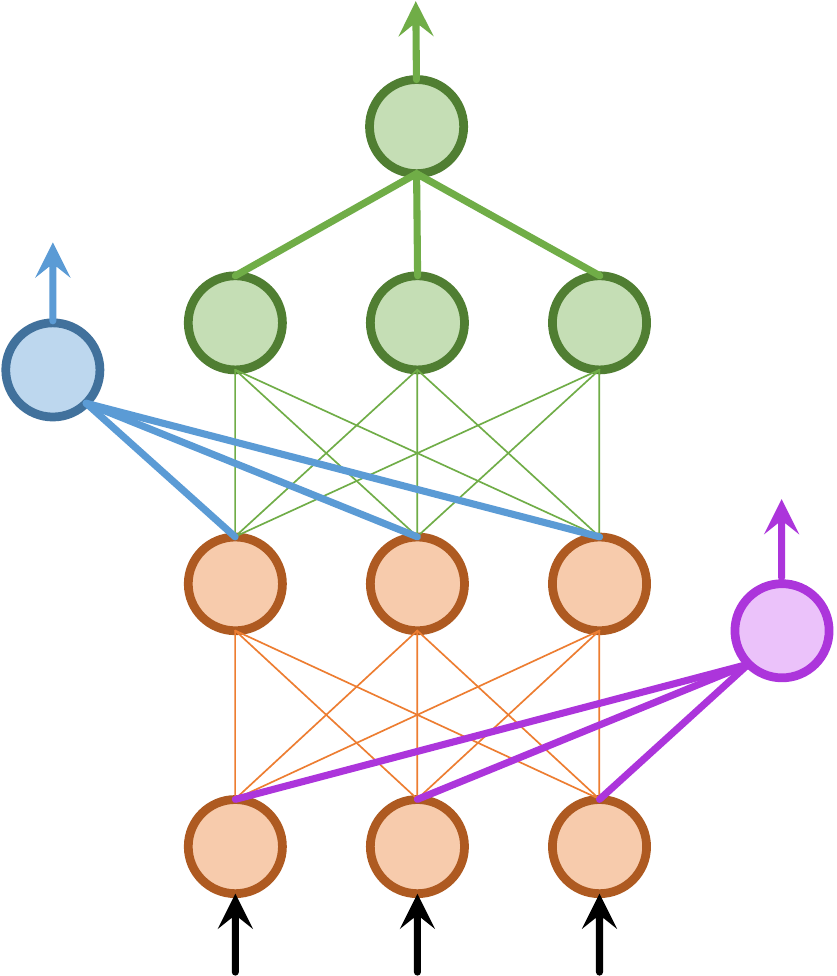}
        \caption{Hierarchical sharing}
        \label{sfig:hier-sharing}
        \end{subfigure}\hfill
    \begin{subfigure}{0.45\linewidth}
        \centering
        \includegraphics[scale=.36]{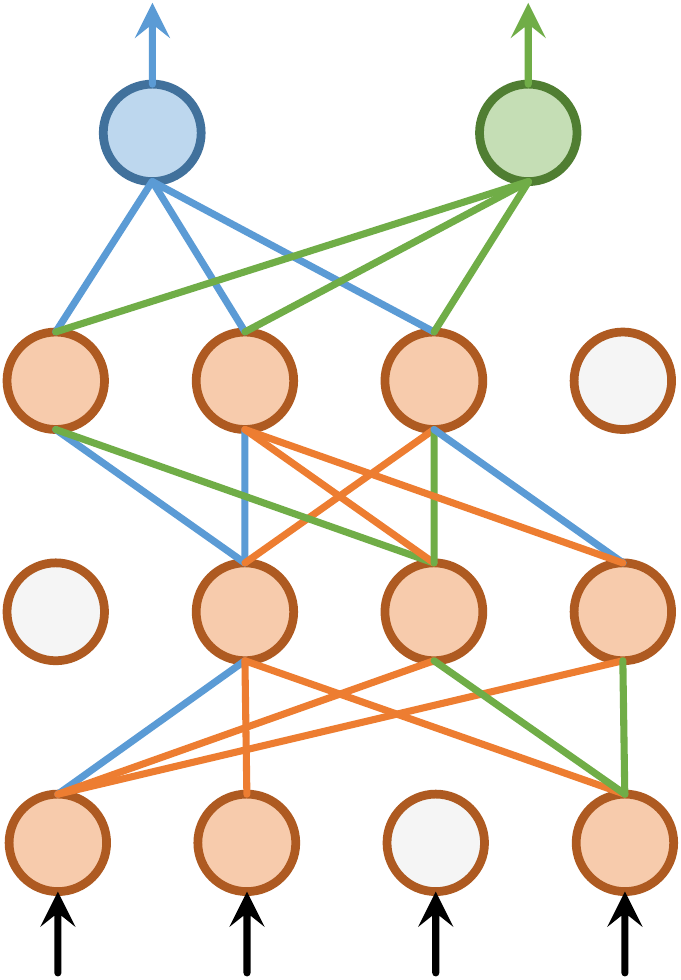}
        \caption{Sparse sharing}
        \label{sfig:sparse-sharing}
        \end{subfigure}\hfill
	\caption{Parameter sharing mechanisms. \includegraphics[scale=0.25]{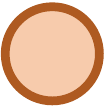} and \textcolor[rgb]{0.68,0.35,0.13}{\textbf{---}} represent shared neurons and weights respectively. \includegraphics[scale=0.25]{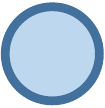}, \includegraphics[scale=0.25]{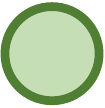}, \includegraphics[scale=0.25]{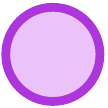} represent task-specific neurons. \textcolor[rgb]{0.36,0.61,0.84}{\textbf{---}}, \textcolor[rgb]{0.44,0.68,0.28}{\textbf{---}}, \textcolor[rgb]{0.67,0.21,0.86}{\textbf{---}} denote task-specific weights for three different tasks.}
	\label{fig:param-sharing}
\end{figure}

Existing work often focuses on knowledge sharing across tasks, which is typically achieved by parameter sharing. Figure \ref{fig:param-sharing}-(a)(b)(c) illustrate three common parameter sharing mechanisms: (a) \emph{hard sharing} approaches stack the task-specific layers on the top of the shared layers~\cite{DBLP:conf/icml/CollobertW08,DBLP:conf/iclr/SubramanianTBP18,DBLP:conf/acl/LiuHCG19}; (b) \emph{soft sharing} approaches allow each task has separate model and parameters, but each model can access the information inside other models \cite{DBLP:conf/cvpr/MisraSGH16,DBLP:conf/ijcai/LiuQH16,DBLP:conf/aaai/RuderBAS19}; (c) \emph{hierarchical sharing} approaches put different tasks at different network layers \cite{DBLP:conf/acl/SogaardG16,DBLP:conf/emnlp/HashimotoXTS17}.

Despite their success, these approaches have some limitations. Hard sharing architectures force all tasks to share the same hidden space, which limits its expressivity and makes it difficult to deal with loosely related tasks. Hierarchical sharing architectures only force part of the model to be shared. Therefore task-specific modules are left with wiggle room to handle heterogeneous tasks. However, designing an effective hierarchy is usually time-consuming. Soft sharing approaches make no assumptions about task relatedness but need to train a model for each task, so are not parameter efficient.

The above limitations motivate us to ask the following question: Does there exist a multi-task sharing architecture that meets the following requirements?
\begin{enumerate}
    \item It is compatible with a wide range of tasks, regardless of whether the tasks are related or not.
    \item It does not depend on manually designing the sharing structure based on characteristic of tasks.
    \item It is parameter efficient.
\end{enumerate}

To answer the above question, we propose a novel parameter sharing mechanism, named \emph{Sparse Sharing}. To obtain a sparse sharing architecture, we start with an over-parameterized neural network, named \emph{Base Network}. Then we extract subnetworks from the base network for each task. Closely related tasks tend to extract similar subnets, thus they can use similar parts of weights, while loosely related or unrelated tasks tend to extract subnets that are different in a wide range. During training, each task only updates the weights of its corresponding subnet. In fact, both hard sharing and hierarchical sharing can be formulated into the framework of sparse sharing, as shown in the latter section. Sparse sharing mechanism is depicted in Figure \ref{fig:param-sharing}(d).

Particularly, we utilize Iterative Magnitude Pruning (IMP) \cite{DBLP:conf/iclr/FrankleC19} to induce sparsity and obtain subnet for each task. Then, these subnets are merged to train all tasks in parallel. Our experiments on sequence labeling tasks demonstrate that sparse sharing meets the requirements mentioned above, and outperforms our single- and multi-task baselines while requiring fewer parameters.

We summarize the our contributions as follows:
\begin{itemize}
    \item We propose a novel parameter-efficient sharing mechanism, Sparse Sharing, for multi-task learning, in which different tasks are partially shared in parameter-level.
        
    \item To induce such a sparse sharing architecture, we propose a simple yet efficient approach based on the Iterative Magnitude Pruning (IMP) \cite{DBLP:conf/iclr/FrankleC19}. Given data of multiple tasks, our approach can automatically learn their sparse sharing architecture, without prior knowledge of task relatedness.
    
    \item Experiments on three sequence labeling tasks demonstrate that sparse sharing architectures achieve consistent improvement compared with single-task learning and existing sharing mechanisms. Moreover, experimental results show that the sparse sharing mechanism helps alleviate the negative transfer, which is a common phenomenon in multi-task learning.
\end{itemize}

\section{Deep Multi-Task Learning}
Multi-task learning (MTL) utilizes the correlation among tasks to improve performance by training tasks in parallel. In this section, we formulate three widely used neural based multi-task learning mechanisms.

Consider $T$ tasks to be learned, with each $t$ associated with a dataset $\mathcal{D}_t=\{x_n^{t}, y_n^{t}\}_{n=1}^{N_t}$ containing $N_t$ samples. Suppose that all tasks have shared layers $\mathcal{E}$ parameterized by $\theta_{\mathcal{E}}=\{\theta_{\mathcal{E}, 1},\dots,\theta_{\mathcal{E}, L}\}$, where $\theta_{\mathcal{E}, l}$ denotes the parameters in the $l$-th network layer. Each task $t$ has its own task-specific layers $\mathcal{F}^t$ parameterized by $\theta_{\mathcal{F}}^t$. Given parameters $\theta=(\theta_\mathcal{E}, \theta_{\mathcal{F}}^1, \dots, \theta_{\mathcal{F}}^T)$ and sample $x_n^t$ of task $t$ as input, the multi-task network predicts
\begin{equation}\label{eq:hard}
    \hat{y}_n^t = \mathcal{F}^t(\mathcal{E}(x_n^t; \theta_\mathcal{E}); \theta_\mathcal{F}^t).
\end{equation}
The parameters of the multi-task model are optimized during joint training with a loss
\begin{equation}
    \mathcal{L}(\theta) = \sum_{t=1}^T\lambda_t\sum_{n=1}^{N_t}\mathcal{L}_t(\hat{y}_n^t, y_n^t),
\end{equation}
where $\mathcal{L}_t(\cdot)$ and $\lambda_t$ are the loss and its weight of task $t$ respectively. In practice, $\lambda_t$ is often considered as a hyper-parameter to be tuned, but can also be treated as a learnable parameter \cite{DBLP:conf/cvpr/KendallGC18}.

Let $\theta_{\mathcal{E}(1:l)} = \{\theta_{\mathcal{E}, 1},\dots,\theta_{\mathcal{E}, l}\}$, then the prediction of task $t$ in hierarchical sharing architecture is
\begin{equation}\label{eq:hier}
    \hat{y}_n^t = \mathcal{F}^t(\mathcal{E}(x_n^t; \theta_{\mathcal{E}(1:l)}); \theta_\mathcal{F}^t),
\end{equation}
if the supervision of task $t$ is put on the $l$-th network layer.

As shown in Eq. (\ref{eq:hard}), task-specific layers can only extract information from the output of the shared layers $\mathcal{E}$, which forces representation of all tasks to be embedded into the same hidden space. In this way, it is hard to learn shared feature for heterogeneous tasks. Hierarchical sharing, as shown in Eq. (\ref{eq:hier}), leaves task-specific modules more space to model heterogeneous tasks by putting supervision at the different layers. However, this does not fundamentally solve the problem. In addition, hierarchical sharing structures are usually hand-crafted, which heavily depends on skills and insights of experts.

For soft sharing models, each task has its own hidden layers as the same as single-task learning. However, during training, each model can access the representations (or other information) produced by the models of other tasks. Obviously, soft sharing is not parameter-efficient because it assigns each task a model.

\section{Learning Sparse Sharing Architectures}
Based on the discussion above, we explore a new multi-task mechanism named \emph{Sparse Sharing}. The architecture of sparse sharing network can be the same as hard sharing, but the parameters in sparse sharing are partially shared.

\begin{figure}[thb]
	\centering
    \includegraphics[width=\columnwidth]{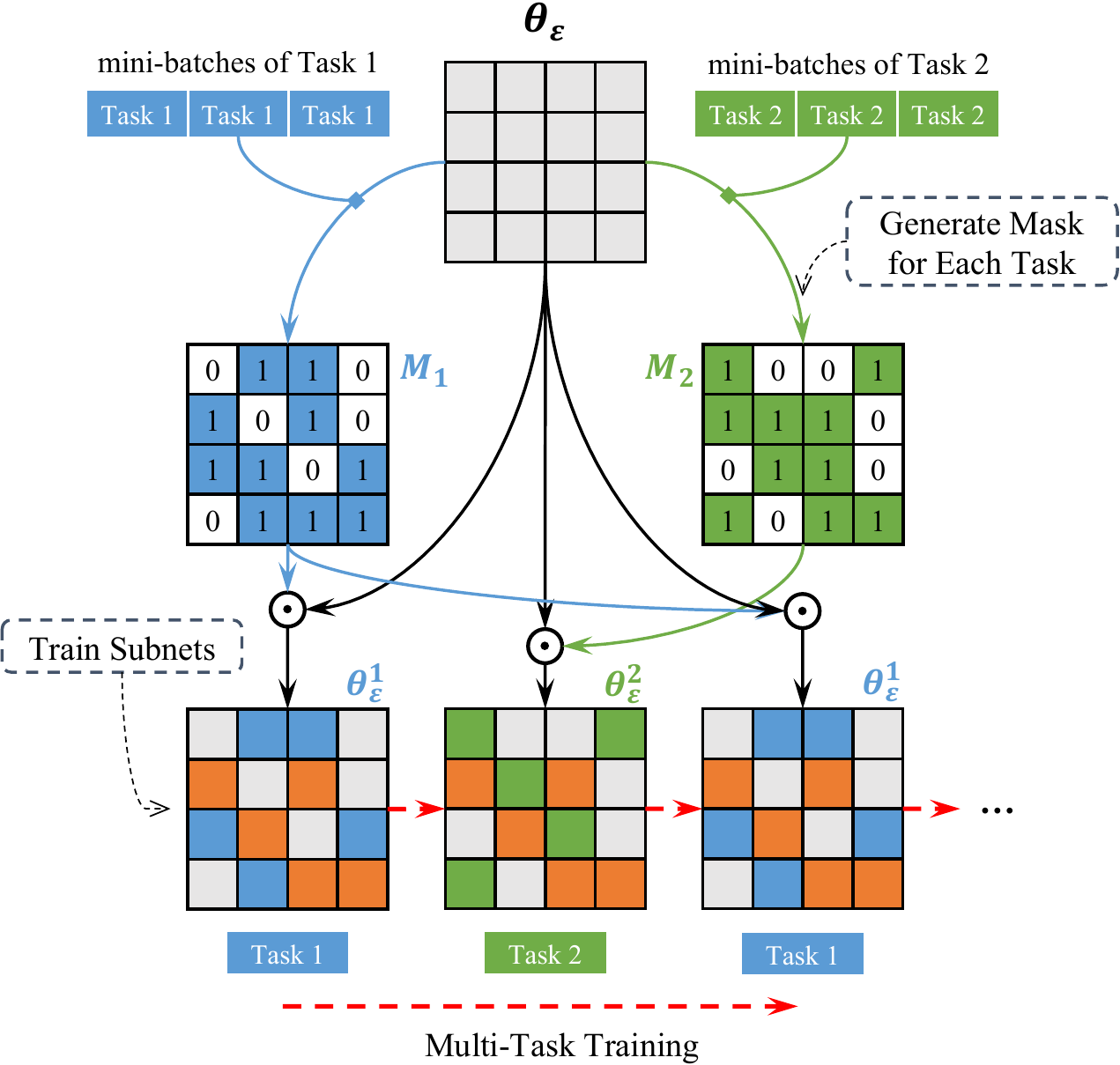}
	\caption{Illustration of our approach to learn sparse sharing architectures. Gray squares are the parameters of a base network. Orange squares represent the shared parameters, while blue and green ones represent private parameters of task 1 and task 2 respectively. }
	\label{fig:frame}
\end{figure}

Sparse sharing starts with an over-parameterized network $\mathcal{E}$, which we call \emph{Base Network}. To handle heterogeneous tasks, we assign each task a different subnet $\mathcal{E}^t$.
Rather than training a model for each task as soft sharing does, we employ a binary mask matrix to select subnet from the base network for each task. Let $M_t\in\{0,1\}^{|\theta_\mathcal{E}|}$ be the mask of task $t$. Then the parameters of subnetwork associated with task $t$ become $M_t\odot \theta_\mathcal{E}$, where $\odot$ denotes element-wise multiplication. In this way, each task can obtain its unique representation $\mathcal{E}^t(x) = \mathcal{E}(x; M_t\odot \theta_\mathcal{E})$. Like other sharing mechanisms, each task has its own task-specific layers $\mathcal{F}^t$.

Note that when $M_t$ takes certain values, sparse sharing can lead to hard sharing and hierarchical sharing. On the one hand, sparse sharing is equivalent to hard sharing if $M_t = \mathbf{1}$. On the other hand, consider a 2-layer base network with parameters $\theta_\mathcal{E} = \{\theta_{\mathcal{E}, 1},\theta_{\mathcal{E}, 2}\}$ and two tasks with $M_1 = \{\mathbf{1}, \mathbf{0}\}$ and $M_2 = \{\mathbf{1}, \mathbf{1}\}$. In this case, the two tasks form a hierarchical sharing architecture.

In particular, our approach can be split into two steps (as shown in Figure \ref{fig:frame}), that is: (1) generating subnets for each task and (2) training subnets in parallel.

\subsection{Generating Subnets for Each Task}
The base model should be over-parameterized so that its hypothesis space is large enough to contain solutions for multiple tasks simultaneously. From the over-parameterized base network, we extract a subnet for each task, whose structure is associated with a hypothesis subspace that suits the given task. In other words, the inductive bias customized to the task is to some extent embedded into the subnet structure. Ideally, tasks with similar inductive bias should be assigned similar parts of parameters. Therefore, subnets corresponding to closely related tasks should share a large number of parameters while subnets corresponding to loosely related tasks should share few parameters.

Recently, some methods are proposed to automatically find network structures for certain tasks, called \emph{Neural Architecture Search} \cite{DBLP:conf/iclr/ZophL17,DBLP:conf/aaai/RealAHL19}. Instead of these methods, we use a simple yet efficient pruning method to find a structure of subnet for each task.

Our method -- Iterative Magnitude Pruning (IMP) -- are inspired by the recently proposed Lottery Ticket Hypothesis \cite{DBLP:conf/iclr/FrankleC19}. The Lottery Ticket Hypothesis states that dense, randomly-initialized, neural networks contain subnetworks (winning tickets) that -- when trained in isolation -- can match test accuracy as the original network. The success of winning tickets also demonstrates the importance of network structures for specific tasks.

Following \citet{DBLP:conf/iclr/FrankleC19}, we employ IMP to generate subnets for tasks. The structures and parameters of subnets are controlled by mask matrices and base network respectively. With the mask $M_t$ for task $t$ and base network $\mathcal{E}$ at hand, the subnetwork for task $t$ can be represented as $\mathcal{E}^t=\mathcal{E}(x;M_t\odot \theta_\mathcal{E})$.

In particular, we perform IMP for each task independently to obtain subnets. Since these subnets are generated from the same base network, the subnet structure of each task encodes how the base network parameters are used for that task. The process of searching for sparse multi-task architecture is detailed in Algorithm \ref{al:imp}. We define the subnet sparsity as $\frac{\|M\|_0}{|\theta|}$.

\floatname{algorithm}{Algorithm}
\begin{algorithm}[hbt]
    \caption{Sparse Sharing Architecture Learning}
    \label{al:imp}
    \begin{algorithmic}[1]
        \Require Base Network $\mathcal{E}$; Pruning rate $\alpha$; Minimal sparsity $S$; Datasets for $T$ tasks $\mathcal{D}_1,\cdots,\mathcal{D}_T$, where $\mathcal{D}_t = \{ x^t_n,y^t_n \}^{N_t}_{n=1}$.
        \State Randomly initialize $\theta_\mathcal{E}$ to $\theta_\mathcal{E}^{(0)}$.
        \For{$t=1\cdots T$}
            \State Initialize mask $M_t^z=\mathbf{1}^{|\theta_\mathcal{E}|}$, where $z=1$.
            \State Train $\mathcal{E}(x; M_t^z\odot \theta_\mathcal{E})$ for $k$ steps with data sampled from $\mathcal{D}_t$, producing network $\mathcal{E}(x; M_t^z\odot \theta_\mathcal{E}^{(k)})$. Let $z \gets z+1$.
            \State Prune $\alpha$ percent of the remaining parameters with the lowest magnitudes from $\theta_\mathcal{E}^{(k)}$. That is, let $M_t^{z}[j]=0$ if $\theta_\mathcal{E}^{(k)}[j]$ is pruned.
            \State If $\frac{\|M_t^z\|_0}{|\theta_\mathcal{E}|}\le S$, the masks for task $t$ are $\{M_t^i\}_{i=1}^z$.
            \State Otherwise, reset $\theta_\mathcal{E}$ to $\theta_\mathcal{E}^{(0)}$ and repeat steps 4-6 iteratively to learn more sparse subnetwork.
        \EndFor
        \State \Return $\{M_1^i\}_{i=1}^z, \{M_2^i\}_{i=1}^z, \cdots \{M_T^i\}_{i=1}^z$.
    \end{algorithmic}
\end{algorithm}

Note that since the pruning is iterative, IMP may generate multiple candidate subnets $\{M_t^i\}_{i=1}^z$ as the pruning proceeds. In practice, we only select one subnet for each task to form the sparse sharing architecture. To address this issue, we adopt a simple principle to select a subnet from multiple candidates. That is, picking the subnet that performs best on the development set. If there are multiple best-performing subnets, we take the subnet with the lowest sparsity.

\subsection{Training Subnets in Parallel}
Finally, we train these subnets with multiple tasks in parallel until convergence. Our multi-task training strategy is similar with \citet{DBLP:conf/icml/CollobertW08}, which is in a stochastic manner by looping over the tasks:
\begin{enumerate}
    \item Select the next task $t$.
    \item Select a random mini-batch for task $t$.
    \item Feed this batch of data into the subnetwork corresponding to task $t$, i.e. $\mathcal{E}(x;M_t\odot\theta_\mathcal{E})$.
    \item Update the subnetwork parameters for this task by taking a gradient step with respect to this mini-batch.
    \item Go to 1.
\end{enumerate}

 When selecting the next task (step 1), we use proportional sampling strategy \cite{DBLP:conf/aaai/SanhWR19}. In proportional sampling, the probability of sampling a task is proportional to the relative size of each dataset compared to the cumulative size of all the datasets.

Although each task only uses its own subnet during training and inference, part of parameters in its subnet are updated by other tasks. In this way, almost every task has shared and private parameters.

\subsection{Multi-Task Warmup}
In practice, we find that randomly initialized base network is sensitive to pruning and data order. To stabilize our approach, we introduce the \emph{Multi-Task Warmup} (MTW) before generating subnets.

More precisely, with randomly initialized base network $\mathcal{E}$ at hand, we first utilize training data of multiple tasks to warm up the base network. Let the initial parameters of $\mathcal{E}$ be $\theta_\mathcal{E}^{(0)}$, we train $\mathcal{E}$ for $w$ steps and the warmuped parameters become $\theta_\mathcal{E}^{(w)}$. When generating subnets, at the end of each pruning iteration (the 7-th line in Algorithm \ref{al:imp}), we reset $\theta_\mathcal{E}$ to $\theta_\mathcal{E}^{(w)}$ instead of $\theta_\mathcal{E}^{(0)}$.

The multi-task warmup (MTW) empirically improves performance in our experiments. Besides, we find that MTW reduces the difference between two subnets generated from the same base network on the same task. Similar phenomenon is obtained by \citet{frankle2019lottery} in single-task settings.

\section{Experiments}
We conduct our experiments on three sequence labeling tasks: Part-of-Speech (POS), Named Entity Recognition (NER) and Chunking.

\subsection{Datasets}
Our experiments are carried out on several widely used sequence labeling datasets, including Penn Treebank(PTB) \cite{DBLP:journals/coling/MarcusSM94}, CoNLL-2000 \cite{DBLP:conf/conll/SangB00}, CoNLL-2003 \cite{DBLP:conf/conll/SangM03} and OntoNotes 5.0 English \cite{DBLP:conf/conll/PradhanMXUZ12}.The datasets are organized in 3 multi-task experiments:
\begin{itemize}
    \item \textbf{Exp1 (CoNLL-2003).}\ \ \ POS, NER, Chunking annotations are simultaneously provided in CoNLL-2003.
    \item \textbf{Exp2 (OntoNotes 5.0).}\ \ \ Also, OntoNotes 5.0 provides annotations for the three tasks. Besides, it contains more data, which comes from multiple domains.
    \item \textbf{Exp3 (PTB + CoNLL-2000 + CoNLL-2003).}\ \ \ Although CoNLL-2003 and OntoNotes provide annotations for multiple tasks, few previous works have reported results for all the three tasks. To compare with the previous multi-task approaches, we follow the multi-task setting of \citet{DBLP:conf/aaai/ChenQLH18}. The PTB POS, CoNLL-2000 Chunking, CoNLL-2003 NER are integrated to construct our multi-task dataset.
\end{itemize}

The statistics of the datasets are summarized in Table \ref{tb:ds}. We use the Wall Street Journal (WSJ) portion of PTB for POS. For OntoNotes, data in \texttt{pt} domain is excluded from our experiments due to its lack of NER annotations. The parse bits in OntoNotes are converted to chunking tags as the same as CoNLL-2003. We use the \texttt{BIOES} tagging scheme for NER and \texttt{BIO2} for Chunking.

\begin{table}[thbp]
\centering
\begin{tabular}{lccc}
\toprule
Datasets & Train & Dev & Test \\ \midrule
PTB & 912,344 & 131,768 & 129,654 \\
CoNLL-2000 & 211,727 & - & 47,377 \\
CoNLL-2003 & 204,567 & 51,578 & 46,666 \\
OntoNotes 5.0 & 1,903,815 & 279,495 & 204,235 \\ \bottomrule
\end{tabular}
\caption{Number of tokens for each dataset.}
\label{tb:ds}
\end{table}

\subsection{Model Settings}
In this subsection, we detail our single- and multi-task baselines. For the sake of comparison, we implement our models with the same architecture.

\textbf{Base Model.}\ \ \ We employ a classical CNN-BiLSTM architecture \cite{DBLP:conf/acl/MaH16} as our base model. For each word, a character-level representation is encoded by CNN with character embedding as input. Then the character-level representation and the word embedding are concatenated to feed into the Bi-directional LSTM \cite{DBLP:journals/neco/HochreiterS97} network. We use 100-dimensional GloVe \cite{DBLP:conf/emnlp/PenningtonSM14} trained on 6 billion words from Wikipedia and web text as our initial word embeddings. Dropout \cite{DBLP:journals/jmlr/SrivastavaHKSS14} is applied after embedding layer and before output layer. We use stochastic gradient descent (SGD) to optimize our models. Main hyper-parameters are summarized in Table \ref{tab:hyper}.

\begin{table}[htb]
\centering
\begin{tabular}{lc}
\toprule
\multicolumn{2}{c}{Hyper-parameters} \\ \midrule
Embedding dimension & 100 \\
Convolution width & 3 \\
CNN output size & 30 \\
LSTM hidden size & 200 \\
Learning rate & 0.1 \\
Dropout & 0.5 \\
Mini-batch size & 10 \\ \bottomrule
\end{tabular}
\caption{Hyper-parameters used in our experiments}
\label{tab:hyper}
\end{table}

\begin{table*}[tbp]
\centering
\begin{tabular}{lccccccc}
	\toprule
	& \multicolumn{2}{c}{POS} & \multicolumn{2}{c}{NER} & \multicolumn{2}{c}{Chunking} &  \\
	\multirow{-2}{*}{Systems} & Test Acc. & $\Delta$ & Test F1 & $\Delta$ & Test F1 & $\Delta$ & \multirow{-2}{*}{\# Params} \\ \midrule[.5pt]
	\multicolumn{8}{c}{Exp1: CoNLL-2003} \\ \midrule[.5pt]
	Single task & 95.09 & - & 89.36 & - & 89.92 & - & 1602k \\
	Single task (subnet) & 95.11 & $+0.02$ & 89.39 & $+0.03$ & 89.96 & $+0.04$ & 811k \\
	Hard sharing & 95.34 & $+0.25$ & 88.68 & {\cellcolor[HTML]{D6D6D6} $-0.68$} & 90.92 & $+1.00$ & 534k \\
	Soft sharing & 95.16 & $+0.07$ & 89.35 & {\cellcolor[HTML]{D6D6D6} $-0.01$} & 90.71 & $+0.79$ & 1596k \\
	Hierarchical sharing & 95.09 & $+0.00$ & 89.30 & {\cellcolor[HTML]{D6D6D6} $-0.06$} & 90.89 & $+0.97$ & 1497k \\
	\textbf{Sparse sharing (ours)} & \textbf{95.56} & $+0.47$ & \textbf{90.35} & $+0.99$ & \textbf{91.55} & $+1.63$ & \textbf{396k} \\ \midrule[.5pt]
	\multicolumn{8}{c}{Exp2: OntoNotes 5.0} \\ \midrule[.5pt]
	Single task & 97.40 & - & 82.72 & - & 95.21 & - & 4491k \\
	Single task (subnet) & 97.42 & $+0.02$ & 82.94 & $+0.22$ & 95.28 & $+0.07$ & 1459k \\
	Hard sharing & 97.46 & $+0.06$ & 82.95 & $+0.23$ & 95.52 & $+0.31$ & 1497k \\
	Soft sharing & 97.34 & {\cellcolor[HTML]{D6D6D6} $-0.06$} & 81.93 & {\cellcolor[HTML]{D6D6D6} $-0.79$} & 95.29 & $+0.08$ & 4485k \\
	Hierarchical sharing & 97.22 & {\cellcolor[HTML]{D6D6D6} $-0.18$} & 82.81 & $+0.09$ & 95.53 & $+0.32$ & 1497k \\
	\textbf{Sparse sharing (ours)} & \textbf{97.54} & $+0.14$ & \textbf{83.42} & $+0.70$ & \textbf{95.56} & $+0.35$ & \textbf{662k} \\ \bottomrule
\end{tabular}
\caption{Experimental results of Exp1 and Exp2. $\Delta$ denotes the improvement compared with single task baselines. We report accuracy(\%) for POS and F1 score(\%) for NER and Chunking. The best scores are in bold. The performance deterioration due to negative transfer is in gray. The embeddings and output layer are excluded when counting parameters. For the sake of simplicity and focusing the comparison of different shared mechanisms, we just use a simple fully-connected output layer as decoder in Exp1 and Exp2.
}
\label{tb:main_results}
\end{table*}

\textbf{Multi-Task Baselines.}\ \ \ We compare the proposed sparse sharing mechanism with three existing mechanisms: hard sharing, soft sharing and hierarchical sharing. The base model described above is used as building block to construct our multi-task baselines.

In Exp1, we use 2-layer BiLSTM for hierarchical sharing and 1-layer BiLSTM for other settings. In Exp2, we use 2-layer BiLSTM in all experiments.
For the sake of simplicity and focusing the comparison of different shared mechanisms, we just use a simple fully-connected output layer as decoder in Exp1 and Exp2.

In Exp3, for fair comparison with previous methods, we employ CRF \cite{DBLP:conf/icml/LaffertyMP01} instead of fully-connected layer as task-specific layer. For hierarchical sharing, following the model setting of \citet{DBLP:conf/acl/SogaardG16}, we put POS task supervision on the inner (1st) layer, put NER and Chunking task supervision on the outer (2nd) layer. For soft sharing, we implement the Cross-stitch network \cite{DBLP:conf/cvpr/MisraSGH16} as our baseline.

\subsection{Details in Subnets Generation}
Note that several hyper-parameters control the generation of subnets, such as the number of multi-task warmup (MTW) steps $w$, pruning rate $\alpha$, etc.

\begin{table}[htbp]
\centering
\begin{tabular}{lccc}
\toprule
     & POS     & NER     & Chunk   \\ \midrule
Exp1 & 50.12\% & 56.23\% & 44.67\% \\
Exp2 & 31.62\% & 25.12\% & 39.81\% \\
Exp3 & 56.23\% & 56.23\% & 56.23\% \\
\bottomrule
\end{tabular}
\caption{The sparsity (percent of remaining parameters) of our selected subnets.}
\label{tb:sparsity}
\end{table}

\begin{figure}[ht]
    \centering
    \includegraphics[width=.9\columnwidth]{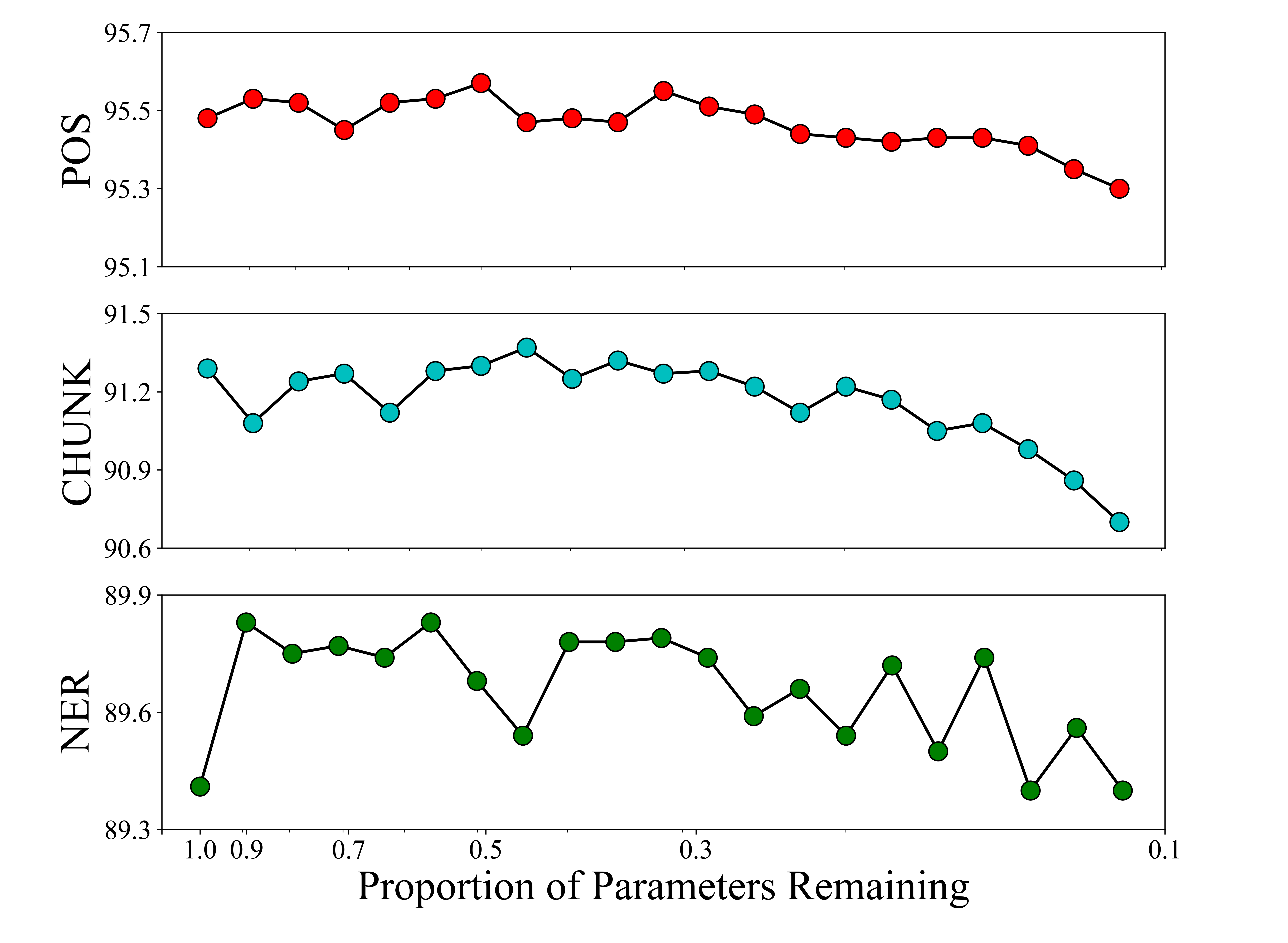}
    \caption{Performance on CoNLL-2003 development set with iteratively pruning. Each point represents a subnet.}
    \label{fig:pruning}
\end{figure}

In all of our experiments, we use global pruning with $\alpha=0.1$. Word embeddings are excluded from pruning. Besides, we set the MTW steps $w=20, 10, 10$ epochs in Exp1, Exp2, Exp3 respectively. As mentioned above, the IMP algorithm produces multiple candidate subnets (with different sparsity) for each task, as shown in Figure \ref{fig:pruning}. It is time-consuming to test all the combinations of these subnets. When selecting subnets, we simply choose the one that performs best on the development set. In fact, we find that a wide range of combinations are effective. The final sparsity of our selected subnets are listed in Table \ref{tb:sparsity}.

\subsection{Main Results}
In this subsection, we present our main results on each experiment (Exp1, Exp2 and Exp3). Our proposed sparse sharing mechanism consistently outperforms our single-task and multi-task baselines while requires fewer parameters. Table \ref{tb:main_results} summarizes the results of Exp1 and Exp2. Table \ref{tab:exp3} shows experimental results on Exp3.

In Exp1 and Exp2, though CoNLL-2003 and OntoNotes are well-suited for multi-task experiments, few previous works have reported the results on all the three tasks simultaneously. \citet{DBLP:conf/aaai/RuderBAS19} conduct their multi-task experiments on OntoNotes, but they only report the accuracy for each task instead of F1 score. Besides, the tasks they use are different from ours. To be able to compare with other parameter sharing mechanisms, we implement our single- and multi-task baselines using the same architecture (CNN-BiLSTM) and hyper-parameters.

To further compare our work with the various single- and multi-task models, we follow the experimental setting of \citet{DBLP:conf/aaai/ChenQLH18} and conduct Exp3. Baselines in Exp3 are implemented in the previous works. For fair comparison, we use CRFs as task-specific layers in Exp3.

\begin{table}[tb]
\centering
\begin{tabular}{lccc}
\toprule
 Systems & POS & NER & Chunk. \\ \midrule
 \textit{Single Task Models:} & & & \\
\citet{DBLP:journals/jmlr/CollobertWBKKK11} & 97.29 & 89.59 & \textbf{94.32} \\
\citet{huang2015bidirectional} & 97.25 & 89.91 & 93.67 \\
\citet{DBLP:conf/aaai/ChenQLH18} & \textbf{97.30} & \textbf{90.08} & 93.71 \\ \midrule
 \textit{Multi-Task Models: }& & & \\
Hard sharing\dag & 97.23 & 90.38 & 94.32 \\
Meta-MTL-LSTM\dag & 97.45 & 90.72 & 95.11 \\
 \textbf{Sparse sharing (ours)} & \textbf{97.57} & \textbf{90.87} & \textbf{95.26} \\ \bottomrule
\end{tabular}
\caption{Experimental results on Exp3. POS, NER and Chunking tasks come from PTB, CoNLL-2003, CoNLL-2000 respectively. \dag\ denotes the model is implemented by \citet{DBLP:conf/aaai/ChenQLH18}. All listed models are equipped with CRF.}
\label{tab:exp3}
\end{table}

\textbf{Where does the benefit come from?}\ \ To figure out where the benefit of sparse sharing comes from, we evaluate the performance of generated subnets on their corresponding tasks in the single-task setting. As shown in the second row of Table \ref{tb:main_results}, each tasks' subnet does not significantly improve the performance on that task. Thus the benefit comes from shared knowledge in other tasks, rather than pruning.

\textbf{Avoiding negative transfer.}\ \ Our experimental results show that multi-task learning does not always yield improvements. Sometimes it even hurts the performance on some tasks, as shown in Table \ref{tb:main_results}, which is called \textit{negative transfer}. Surprisingly, we find that sparse sharing mechanism is helpful to avoid negative transfer.

\section{Analysis and Discussions}
In this section, we further analyze the ability of sparse sharing on negative transfer, task interaction and sparsity. At last, an ablation study about multi-task warmup is conducted.

\subsection{About Negative Transfer}
Negative effects usually occurs when there are unrelated tasks. To further analyze the impact of negative transfer for our model, we construct two loosely related tasks. One task is the real CoNLL-2003 NER task. The other task, Position Prediction (PP), is synthetic. The PP task is to predict each token's position in the sentence. The PP task annotation is collected from CoNLL-2003. Thus, the PP task and CoNLL-2003 NER are two loosely related tasks and form a multi-task setting.

We employ a 1-layer BiLSTM with 50-dimensional hidden size as shared module and a fully-connected layer as task-specific layer. Our word embeddings are in 50 dimensions and randomly initialized. Our experimental results are shown in Table \ref{tb:task_rel}. Note that $\Delta$ is defined as the increase of performance compared with single-task models. The synthetic experiment shows that hard sharing suffers from negative transfer, while sparse sharing does not.

\begin{table}[htbp]
\centering
\begin{tabular}{l|cc|cc}
\toprule
 & NER & $\Delta$ & PP & $\Delta$ \\ \midrule
Single task & 71.05 & - & 99.21 & - \\
Hard sharing & {\cellcolor[HTML]{D6D6D6}61.62} & {\cellcolor[HTML]{D6D6D6} $-9.43$} & 99.50 & $+0.29$ \\
Sparse sharing & 71.46 & $+0.41$ & 99.45 & $+0.24$ \\ \bottomrule
\end{tabular}
\caption{Results of the synthetic experiment. The performance deterioration due to negative transfer is in gray.}
\label{tb:task_rel}
\end{table}

\subsection{About Task Relatedness}

Furthermore, we quantitatively discuss the task relatedness and its relationship with multi-task benefit. We provide a novel perspective to analyze task relatedness.
We define the Overlap Ratio (OR) among mask matrices as the zero-norm of their intersection divided by the zero-norm of their union:
\begin{equation}\label{eq:or}
    \mathrm{OR}(M_1, M_2, \cdots, M_T) = \frac{\| \cap_{t=1}^TM_t\|_0}{\|\cup_{t=1}^TM_t\|_0}.
\end{equation}
Each mask matrix is associated with a subnet. On the one hand, OR reflects the similarity among subnets. On the other hand, OR reflects the degree of sharing among tasks.

We group the three tasks on CoNLL-2003 into pairs and evaluate the performance of sparse and hard sharing on these task pairs. As shown in Table \ref{tb:task_or}, we find that the larger the mask OR, which means the more correlated the two tasks are, the smaller the improvement of sparse sharing compared with hard sharing. This suggests that when tasks are closely related, hard sharing is still an efficient parameter sharing mechanism.

\begin{table}[!htbp]
\centering
\begin{tabular}{ccc}
\toprule
Task Pairs & Mask $\mathrm{OR}$ & $\Delta(S^2-HS)$ \\ \midrule
POS \& NER & 0.18 & 0.4 \\
NER \& Chunking & 0.20 & 0.34 \\
POS \& Chunking & 0.50 & 0.05 \\ \bottomrule
\end{tabular}
\caption{Mask Overlap Ratio ($\mathrm{OR}$) and the improvement for sparse sharing ($S^2$) compared to hard sharing ($HS$) of tasks on CoNLL-2003. The improvement is calculated using the average performance on the test set.}
\label{tb:task_or}
\end{table}

\subsection{About Sparsity}
\begin{figure}[!htbp]
    \centering
    \includegraphics[width=.9\columnwidth]{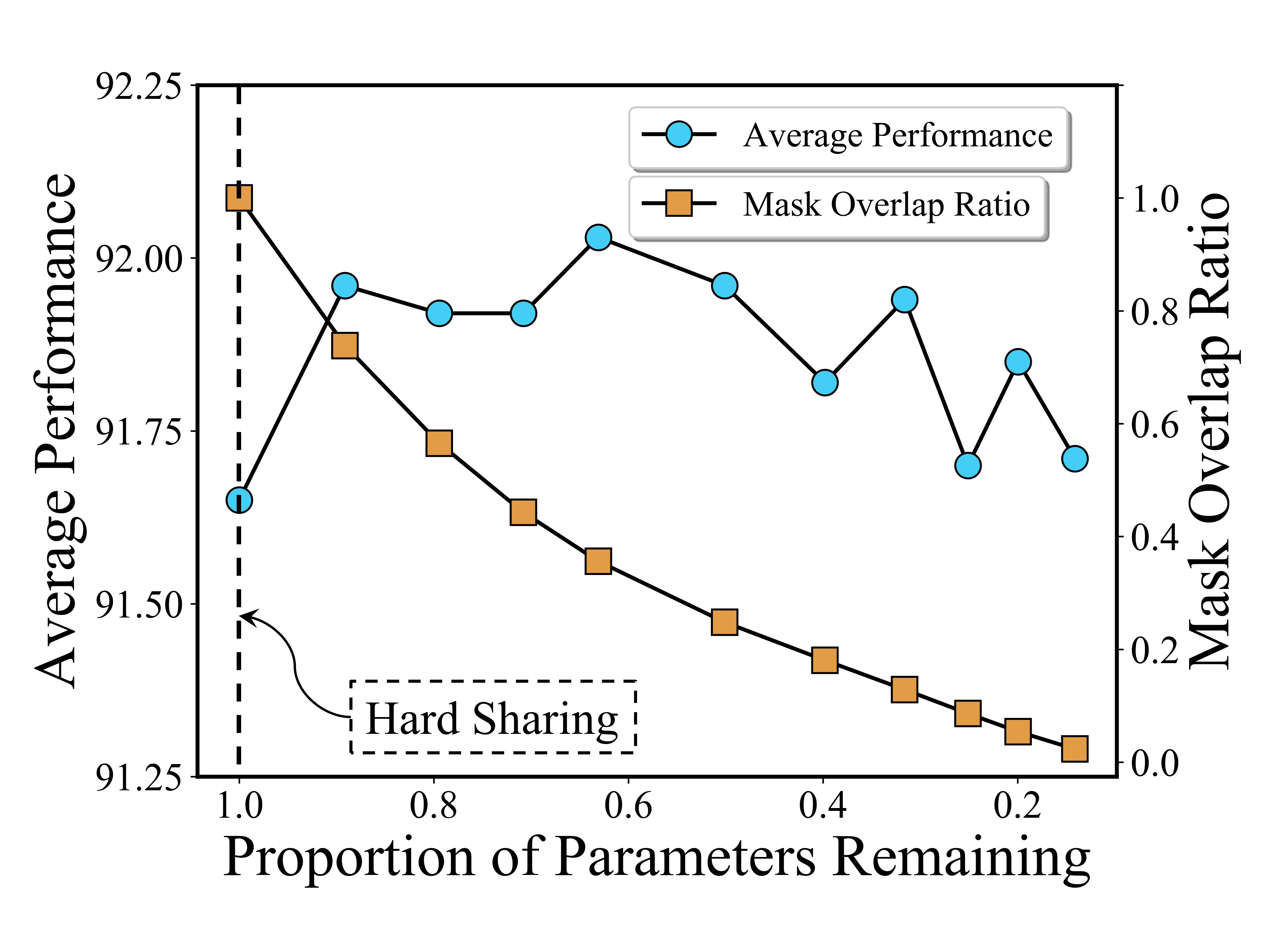}
    \caption{Average test performance and mask OR with different sparsity combinations.}
    \label{fig:avg_or}
\end{figure}

In addition, we evaluate various combinations of subnets with different sparsity. For the sake of simplicity, we select subnet for each task with the same sparsity to construct the sparse sharing architecture. Figure \ref{fig:avg_or} plots the average test performance and mask OR with different sparsity combinations. Our evaluation is carried out on CoNLL-2003.

When $M_1=M_2=M_3=\mathbf{1}$, the sparsity of subnets for the three tasks is 100\%, and the mask OR takes 1. In this case, sparse sharing is equivalent to hard sharing. With the decrease of sparsity, the mask OR also decreases while the average performance fluctuates.

\subsection{About Multi-Task Warmup}
At last, we conduct an ablation study about multi-task warmup (MTW). The performance achieved by the model with and without MTW are listed in Table \ref{tb:warmup}. In most instances, the model with MTW achieves better performance.

\begin{table}[htb]
\centering
\begin{tabular}{llll}
\toprule
 & POS & NER & Chunking \\ \midrule
\multicolumn{4}{c}{CoNLL-2003} \\ \midrule
Sparse sharing & 95.56 & 90.35 & 91.55 \\
\ \ \ $-$ MTW & 95.36 & 89.62 & 91.04 \\ \midrule
\multicolumn{4}{c}{OntoNotes 5.0} \\ \midrule
Sparse sharing & 97.54 & 83.42 & 95.56 \\
\ \ \ $-$ MTW & 97.53 & 81.15 & 95.48 \\ \bottomrule
\end{tabular}
\caption{Test accuracy (for POS) and F1 (for NER and Chunking) on CoNLL-2003 and OntoNotes 5.0. MTW: Multi-Task Warmup.}
\label{tb:warmup}
\end{table}

\section{Related Work}
There are two lines of research related to our work -- deep multi-task learning and sparse neural networks.

Neural based multi-task learning approaches can be roughly grouped into three folds: (1) hard sharing, (2) hierarchical sharing, and (3) soft sharing. Hard sharing finds for a representation that is preferred by as many tasks as possible \cite{DBLP:journals/ml/Caruana97}. In spite of its simple and parameter-efficient, it is only guaranteed to work for closely related tasks \cite{DBLP:journals/jair/Baxter00}. Hierarchical sharing approaches put different level of tasks on the different network layer \cite{DBLP:conf/acl/SogaardG16,DBLP:conf/emnlp/HashimotoXTS17}, which to some extent, relax the constraint about task relatedness. However, the hierarchy of tasks is usually designed by hand through the skill and insights of experts. Besides, tasks can hurt each other when they are embedded into the same hidden space \cite{DBLP:conf/eacl/PlankA17}. To mitigate negative transfer, soft sharing is proposed and achieves success in computer vision \cite{DBLP:conf/cvpr/MisraSGH16} and natural language processing \cite{DBLP:conf/aaai/LiuFDQC19,DBLP:conf/aaai/RuderBAS19}. In spite of its flexible, soft sharing allows each task to have its separate parameters, thus is parameter inefficient. Different from these work, sparse sharing is a fine-grained parameter sharing strategy that is flexible to handle heterogeneous tasks and parameter efficient.

Our approach is also inspired by the sparsity of networks, especially the Lottery Ticket Hypothesis \cite{DBLP:conf/iclr/FrankleC19}. \citet{DBLP:conf/iclr/FrankleC19} finds that a subnet (winning ticket) -- that can reach test accuracy comparable to the original network -- can be produced through Iterative Magnitude Pruning (IMP). Further, \citet{frankle2019lottery} introduce late resetting to stabilize the lottery ticket hypothesis at scale. Besides, \citet{yu2019playing} confirms that winning ticket initializations exist in LSTM and NLP tasks. Our experiments also demonstrate that winning tickets do exist in sequence labeling tasks. In addition, it is worth noticing that sparse sharing architecture is also possible to be learned using variational dropout \cite{DBLP:conf/icml/MolchanovAV17}, $l_0$ regularization \cite{louizos2017learning} or other pruning techniques.

\section{Conclusion}
Most existing neural-based multi-task learning models are done with parameter sharing, e.g. hard sharing, soft sharing, hierarchical sharing etc. These sharing mechanisms have some inevitable limitations: (1) hard sharing struggles with heterogeneous tasks, (2) soft sharing is parameter-inefficient, (3) hierarchical sharing depends on manually design. To alleviate the limitations, we propose a novel parameter sharing mechanism, named Sparse Sharing. The parameters in sparse sharing architectures are partially shared across tasks, which makes it flexible to handle loosely related tasks. 

To induce such architectures, we propose a simple yet efficient approach that can automatically extract subnets for each task. The obtained subnets are overlapped and trained in parallel. Our experimental results show that sparse sharing architectures achieve consistent improvement while requiring fewer parameters. Besides, our synthetic experiment shows that sparse sharing avoids negative transfer even when tasks are unrelated.

\section{Acknowledgments}
We would like to thank the anonymous reviewers for their valuable comments. This work was supported by the National Key Research and Development Program of China (No. 2018YFC0831103), National Natural Science Foundation of China (No. 61672162), Shanghai Municipal Science and Technology Major Project (No. 2018SHZDZX01) and ZJLab.

\bibliographystyle{aaai}
\bibliography{3353_ref}

\begin{thebibliography}{}

\bibitem[\protect\citeauthoryear{Baxter}{2000}]{DBLP:journals/jair/Baxter00}
Baxter, J.
\newblock 2000.
\newblock A model of inductive bias learning.
\newblock {\em J. Artif. Intell. Res.} 12:149--198.

\bibitem[\protect\citeauthoryear{Caruana}{1997}]{DBLP:journals/ml/Caruana97}
Caruana, R.
\newblock 1997.
\newblock Multitask learning.
\newblock {\em Machine Learning} 28(1):41--75.

\bibitem[\protect\citeauthoryear{Chen \bgroup et al\mbox.\egroup
  }{2018}]{DBLP:conf/aaai/ChenQLH18}
Chen, J.; Qiu, X.; Liu, P.; and Huang, X.
\newblock 2018.
\newblock Meta multi-task learning for sequence modeling.
\newblock In {\em {AAAI}},  5070--5077.

\bibitem[\protect\citeauthoryear{Collobert and
  Weston}{2008}]{DBLP:conf/icml/CollobertW08}
Collobert, R., and Weston, J.
\newblock 2008.
\newblock A unified architecture for natural language processing: deep neural
  networks with multitask learning.
\newblock In {\em {ICML}},  160--167.

\bibitem[\protect\citeauthoryear{Collobert \bgroup et al\mbox.\egroup
  }{2011}]{DBLP:journals/jmlr/CollobertWBKKK11}
Collobert, R.; Weston, J.; Bottou, L.; Karlen, M.; Kavukcuoglu, K.; and Kuksa,
  P.~P.
\newblock 2011.
\newblock Natural language processing (almost) from scratch.
\newblock {\em J. Mach. Learn. Res.} 12:2493--2537.

\bibitem[\protect\citeauthoryear{Frankle and
  Carbin}{2019}]{DBLP:conf/iclr/FrankleC19}
Frankle, J., and Carbin, M.
\newblock 2019.
\newblock The lottery ticket hypothesis: Finding sparse, trainable neural
  networks.
\newblock In {\em {ICLR}}.

\bibitem[\protect\citeauthoryear{Frankle \bgroup et al\mbox.\egroup
  }{2019}]{frankle2019lottery}
Frankle, J.; Dziugaite, G.~K.; Roy, D.~M.; and Carbin, M.
\newblock 2019.
\newblock The lottery ticket hypothesis at scale.
\newblock {\em arXiv preprint arXiv:1903.01611}.

\bibitem[\protect\citeauthoryear{Hashimoto \bgroup et al\mbox.\egroup
  }{2017}]{DBLP:conf/emnlp/HashimotoXTS17}
Hashimoto, K.; Xiong, C.; Tsuruoka, Y.; and Socher, R.
\newblock 2017.
\newblock A joint many-task model: Growing a neural network for multiple {NLP}
  tasks.
\newblock In {\em {EMNLP}},  1923--1933.

\bibitem[\protect\citeauthoryear{Hochreiter and
  Schmidhuber}{1997}]{DBLP:journals/neco/HochreiterS97}
Hochreiter, S., and Schmidhuber, J.
\newblock 1997.
\newblock Long short-term memory.
\newblock {\em Neural Computation} 9(8):1735--1780.

\bibitem[\protect\citeauthoryear{Huang, Xu, and
  Yu}{2015}]{huang2015bidirectional}
Huang, Z.; Xu, W.; and Yu, K.
\newblock 2015.
\newblock Bidirectional lstm-crf models for sequence tagging.
\newblock {\em arXiv preprint arXiv:1508.01991}.

\bibitem[\protect\citeauthoryear{Kendall, Gal, and
  Cipolla}{2018}]{DBLP:conf/cvpr/KendallGC18}
Kendall, A.; Gal, Y.; and Cipolla, R.
\newblock 2018.
\newblock Multi-task learning using uncertainty to weigh losses for scene
  geometry and semantics.
\newblock In {\em {CVPR}},  7482--7491.

\bibitem[\protect\citeauthoryear{Lafferty, McCallum, and
  Pereira}{2001}]{DBLP:conf/icml/LaffertyMP01}
Lafferty, J.~D.; McCallum, A.; and Pereira, F. C.~N.
\newblock 2001.
\newblock Conditional random fields: Probabilistic models for segmenting and
  labeling sequence data.
\newblock In {\em {ICML}},  282--289.

\bibitem[\protect\citeauthoryear{Liu \bgroup et al\mbox.\egroup
  }{2019a}]{DBLP:conf/aaai/LiuFDQC19}
Liu, P.; Fu, J.; Dong, Y.; Qiu, X.; and Cheung, J. C.~K.
\newblock 2019a.
\newblock Learning multi-task communication with message passing for sequence
  learning.
\newblock In {\em {AAAI}},  4360--4367.

\bibitem[\protect\citeauthoryear{Liu \bgroup et al\mbox.\egroup
  }{2019b}]{DBLP:conf/acl/LiuHCG19}
Liu, X.; He, P.; Chen, W.; and Gao, J.
\newblock 2019b.
\newblock Multi-task deep neural networks for natural language understanding.
\newblock In {\em {ACL}},  4487--4496.

\bibitem[\protect\citeauthoryear{Liu, Qiu, and
  Huang}{2016}]{DBLP:conf/ijcai/LiuQH16}
Liu, P.; Qiu, X.; and Huang, X.
\newblock 2016.
\newblock Recurrent neural network for text classification with multi-task
  learning.
\newblock In {\em {IJCAI}},  2873--2879.

\bibitem[\protect\citeauthoryear{Liu, Qiu, and
  Huang}{2017}]{DBLP:conf/acl/LiuQH17}
Liu, P.; Qiu, X.; and Huang, X.
\newblock 2017.
\newblock Adversarial multi-task learning for text classification.
\newblock In {\em {ACL}},  1--10.

\bibitem[\protect\citeauthoryear{Louizos, Welling, and
  Kingma}{2017}]{louizos2017learning}
Louizos, C.; Welling, M.; and Kingma, D.~P.
\newblock 2017.
\newblock Learning sparse neural networks through $ l\_0 $ regularization.
\newblock {\em arXiv preprint arXiv:1712.01312}.

\bibitem[\protect\citeauthoryear{Luong \bgroup et al\mbox.\egroup
  }{2016}]{DBLP:journals/corr/LuongLSVK15}
Luong, M.; Le, Q.~V.; Sutskever, I.; Vinyals, O.; and Kaiser, L.
\newblock 2016.
\newblock Multi-task sequence to sequence learning.
\newblock In {\em {ICLR}}.

\bibitem[\protect\citeauthoryear{Ma and Hovy}{2016}]{DBLP:conf/acl/MaH16}
Ma, X., and Hovy, E.~H.
\newblock 2016.
\newblock End-to-end sequence labeling via bi-directional lstm-cnns-crf.
\newblock In {\em {ACL}}.

\bibitem[\protect\citeauthoryear{Marcus, Santorini, and
  Marcinkiewicz}{1993}]{DBLP:journals/coling/MarcusSM94}
Marcus, M.~P.; Santorini, B.; and Marcinkiewicz, M.~A.
\newblock 1993.
\newblock Building a large annotated corpus of english: The penn treebank.
\newblock {\em Computational Linguistics} 19(2):313--330.

\bibitem[\protect\citeauthoryear{Misra \bgroup et al\mbox.\egroup
  }{2016}]{DBLP:conf/cvpr/MisraSGH16}
Misra, I.; Shrivastava, A.; Gupta, A.; and Hebert, M.
\newblock 2016.
\newblock Cross-stitch networks for multi-task learning.
\newblock In {\em {CVPR}},  3994--4003.

\bibitem[\protect\citeauthoryear{Molchanov, Ashukha, and
  Vetrov}{2017}]{DBLP:conf/icml/MolchanovAV17}
Molchanov, D.; Ashukha, A.; and Vetrov, D.~P.
\newblock 2017.
\newblock Variational dropout sparsifies deep neural networks.
\newblock In {\em {ICML}},  2498--2507.

\bibitem[\protect\citeauthoryear{Pennington, Socher, and
  Manning}{2014}]{DBLP:conf/emnlp/PenningtonSM14}
Pennington, J.; Socher, R.; and Manning, C.~D.
\newblock 2014.
\newblock Glove: Global vectors for word representation.
\newblock In {\em {EMNLP}},  1532--1543.

\bibitem[\protect\citeauthoryear{Plank and
  Alonso}{2017}]{DBLP:conf/eacl/PlankA17}
Plank, B., and Alonso, H.~M.
\newblock 2017.
\newblock When is multitask learning effective? semantic sequence prediction
  under varying data conditions.
\newblock In {\em {EACL}},  44--53.

\bibitem[\protect\citeauthoryear{Pradhan \bgroup et al\mbox.\egroup
  }{2012}]{DBLP:conf/conll/PradhanMXUZ12}
Pradhan, S.; Moschitti, A.; Xue, N.; Uryupina, O.; and Zhang, Y.
\newblock 2012.
\newblock Conll-2012 shared task: Modeling multilingual unrestricted
  coreference in ontonotes.
\newblock In {\em {EMNLP-CoNLL Shared Task}},  1--40.

\bibitem[\protect\citeauthoryear{Real \bgroup et al\mbox.\egroup
  }{2019}]{DBLP:conf/aaai/RealAHL19}
Real, E.; Aggarwal, A.; Huang, Y.; and Le, Q.~V.
\newblock 2019.
\newblock Regularized evolution for image classifier architecture search.
\newblock In {\em {AAAI}},  4780--4789.

\bibitem[\protect\citeauthoryear{Ruder \bgroup et al\mbox.\egroup
  }{2019}]{DBLP:conf/aaai/RuderBAS19}
Ruder, S.; Bingel, J.; Augenstein, I.; and S{\o}gaard, A.
\newblock 2019.
\newblock Latent multi-task architecture learning.
\newblock In {\em {AAAI}},  4822--4829.

\bibitem[\protect\citeauthoryear{Sang and
  Buchholz}{2000}]{DBLP:conf/conll/SangB00}
Sang, E. F. T.~K., and Buchholz, S.
\newblock 2000.
\newblock Introduction to the conll-2000 shared task chunking.
\newblock In {\em {CoNLL-LLL}},  127--132.

\bibitem[\protect\citeauthoryear{Sang and
  Meulder}{2003}]{DBLP:conf/conll/SangM03}
Sang, E. F. T.~K., and Meulder, F.~D.
\newblock 2003.
\newblock Introduction to the conll-2003 shared task: Language-independent
  named entity recognition.
\newblock In {\em {HLT-NAACL}},  142--147.

\bibitem[\protect\citeauthoryear{Sanh, Wolf, and
  Ruder}{2019}]{DBLP:conf/aaai/SanhWR19}
Sanh, V.; Wolf, T.; and Ruder, S.
\newblock 2019.
\newblock A hierarchical multi-task approach for learning embeddings from
  semantic tasks.
\newblock In {\em {AAAI}},  6949--6956.

\bibitem[\protect\citeauthoryear{S{\o}gaard and
  Goldberg}{2016}]{DBLP:conf/acl/SogaardG16}
S{\o}gaard, A., and Goldberg, Y.
\newblock 2016.
\newblock Deep multi-task learning with low level tasks supervised at lower
  layers.
\newblock In {\em {ACL}}.

\bibitem[\protect\citeauthoryear{Srivastava \bgroup et al\mbox.\egroup
  }{2014}]{DBLP:journals/jmlr/SrivastavaHKSS14}
Srivastava, N.; Hinton, G.~E.; Krizhevsky, A.; Sutskever, I.; and
  Salakhutdinov, R.
\newblock 2014.
\newblock Dropout: a simple way to prevent neural networks from overfitting.
\newblock {\em J. Mach. Learn. Res.} 15(1):1929--1958.

\bibitem[\protect\citeauthoryear{Subramanian \bgroup et al\mbox.\egroup
  }{2018}]{DBLP:conf/iclr/SubramanianTBP18}
Subramanian, S.; Trischler, A.; Bengio, Y.; and Pal, C.~J.
\newblock 2018.
\newblock Learning general purpose distributed sentence representations via
  large scale multi-task learning.
\newblock In {\em {ICLR}}.

\bibitem[\protect\citeauthoryear{Yu \bgroup et al\mbox.\egroup
  }{2019}]{yu2019playing}
Yu, H.; Edunov, S.; Tian, Y.; and Morcos, A.~S.
\newblock 2019.
\newblock Playing the lottery with rewards and multiple languages: lottery
  tickets in rl and nlp.
\newblock {\em arXiv preprint arXiv:1906.02768}.

\bibitem[\protect\citeauthoryear{Zamir \bgroup et al\mbox.\egroup
  }{2018}]{DBLP:conf/cvpr/ZamirSSGMS18}
Zamir, A.~R.; Sax, A.; Shen, W.~B.; Guibas, L.~J.; Malik, J.; and Savarese, S.
\newblock 2018.
\newblock Taskonomy: Disentangling task transfer learning.
\newblock In {\em {CVPR}},  3712--3722.

\bibitem[\protect\citeauthoryear{Zoph and Le}{2017}]{DBLP:conf/iclr/ZophL17}
Zoph, B., and Le, Q.~V.
\newblock 2017.
\newblock Neural architecture search with reinforcement learning.
\newblock In {\em {ICLR}}.

\end{thebibliography}
\end{document}